\documentclass[letterpaper, 10 pt, conference]{ieeeconf}  

\IEEEoverridecommandlockouts                              

\usepackage{amsmath} 
\usepackage{amssymb}  
\usepackage{graphicx}
\usepackage{CJKutf8}
\usepackage{tabularx} 
\usepackage{booktabs} 
\usepackage{csquotes} 
\usepackage{adjustbox} 
\usepackage{multirow}
\usepackage{makecell} 
\usepackage[english]{babel}
\usepackage{threeparttablex}
\usepackage[
bookmarks=true,
colorlinks=true,
linkcolor=blue,
urlcolor=blue,
citecolor=blue,
bookmarks=true,
hyperindex=true
]{hyperref}
\usepackage{placeins}
\usepackage{makecell}
\usepackage{xcolor}
\usepackage[backend=biber,maxnames=3,style=ieee,sorting=none,giveninits=true]{biblatex}

\usepackage{calc}

\usepackage{caption}
\usepackage{subcaption}
\allowdisplaybreaks

\AtEveryBibitem{%
    \clearfield{doi}
    \clearfield{url}
    \clearfield{isbn}
    \clearfield{issn}
    \clearfield{eprint}
    \clearfield{location}
    \clearfield{address}
    \clearfield{series}
    \clearfield{location}
}
\title{\LARGE \bf
Driving with Context: Online Map Matching for Complex Roads Using Lane Markings and Scenario Recognition
}

\author{Xin Bi$^{1}$, Zhichao Li$^{1}$, Yuxuan Xia$^{2}$, Panpan Tong$^{1}$, Lijuan Zhang$^{3}$, Yang Chen$^{3}$, and Junsheng Fu$^{3}$
\thanks{
This research was conducted as part of a collaborative project between Zenseact and Tongji University. Zhichao Li accessed the regulated test data in China as an intern to carry out the research work.}
\thanks{$^{1}$Xin Bi, Zhichao Li, and Panpan Tong are with the School of Automotive Studies, Tongji University, Shanghai 201804, China (bixin@tongji.edu.cn; 2231552@tongji.edu.cn; tongpanpan@tongji.edu.cn).}%
\thanks{$^{2}$Yuxuan Xia was a Post-doc with Zenseact, and he is now with the Department of Automation and Perception, Shanghai Jiao Tong University, Shanghai 200240, China. (yuxuan.xia@sjtu.edu.cn).}%
\thanks{$^{3}$Lijuan Zhang, Yang Chen, and Junsheng Fu are with Zenseact, Lindholmspiren 2, Gothenburg 41756, Sweden (lijuan.zhang@zenseact.com; yang.chen@zenseact.com;junsheng.fu@zenseact.com).
}%
}

\begin{document}

\maketitle
\thispagestyle{empty}
\pagestyle{empty}

\begin{abstract}
Accurate online map matching is fundamental to vehicle navigation and the activation of intelligent driving functions. Current online map matching methods are prone to errors in complex road networks, especially in multilevel road area. To address this challenge, we propose an online Standard Definition (SD) map matching method by constructing a Hidden Markov Model (HMM) with multiple probability factors. Our proposed method can achieve accurate map matching even in complex road networks by carefully leveraging lane markings and scenario recognition in the designing of the probability factors. First, the lane markings are generated by a multi-lane tracking method and associated with the SD map using HMM to build an enriched SD map. In areas covered by the enriched SD map, the vehicle can re-localize itself by performing Iterative Closest Point (ICP) registration for the lane markings. Then, the probability factor accounting for the lane marking detection can be obtained using the association probability between adjacent lanes and roads. Second, the driving scenario recognition model is applied to generate the emission probability factor of scenario recognition, which improves the performance of map matching on elevated roads and ordinary urban roads underneath them. We validate our method through extensive road tests in Europe and China, and the experimental results show that our proposed method effectively improves the online map matching accuracy as compared to other existing methods, especially in multilevel road area. Specifically, the experiments show that our proposed method achieves $\mathbf{F_1}$ scores of 98.04\% and 94.60\% on the Zenseact Open Dataset and test data of multilevel road
areas in Shanghai respectively, significantly outperforming benchmark methods. The implementation is available at \href{https://github.com/TRV-Lab/LMSR-OMM}{github.com/TRV-Lab/LMSR-OMM}.

\end{abstract}
\section{INTRODUCTION}
Intelligent driving functions usually require a well defined operational design domain (ODD) \cite{1}, and online map matching enables vehicles to assess whether they are traversing areas and road classes that fall within the ODD for intelligent driving functions. Map matching errors can disrupt navigation systems and erroneously deactivate intelligent driving functions, affecting vehicle trajectory planning and speed control, increasing accident risks.

\begin{figure}
    \centering
    \begin{subfigure}[t]{0.33\textwidth}
        \includegraphics[width=\textwidth]{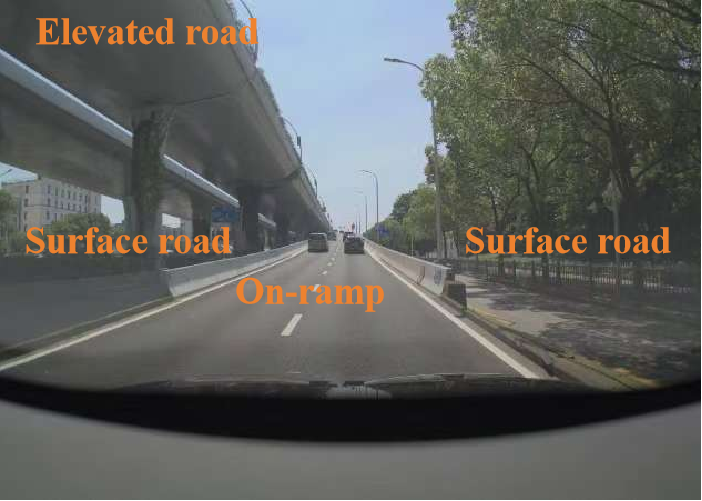}
        \subcaption{Image taken from the front-view camera}\label{fig:sub-1a}
    \end{subfigure}
    \hfill
    \begin{subfigure}[t]{0.144\textwidth}
        \includegraphics[width=\textwidth]{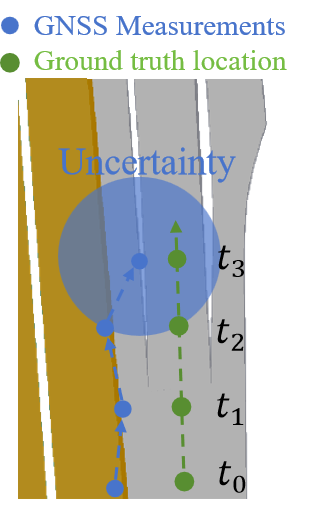}
        \subcaption{An illustration of the ego vehicle trajectory}\label{fig:sub-1b}
    \end{subfigure}
    \caption{
    A challenging scenario in a multilevel road area. In this scenario, the vehicle is driving on the ramp of an elevated road, but the positioning drifts to either the elevated road or the adjacent surface road, making it difficult to achieve accurate map matching.
    }\label{fig:1}
\end{figure}

Existing online map matching methods have inferior performance in complex scenarios, as illustrated in Fig. \ref{fig:1}. There are three main challenges in such scenarios: positioning errors, Standard Definition (SD) map inaccuracies and complex road networks. First, vehicle positioning errors can reach tens of meters in urban canyons \cite{10433884}, significantly reducing map matching accuracy. Second, SD map lacks lane-level details and cannot provide high-precision road network geometric data. This leads to the inability to provide the exact boundary geometry of road splits on the map. In addition, the lack of road height information in certain countries makes it not always possible to assist map matching in multilevel road area. Third, incorrect matching is likely to occur at road splits e.g. expressway ramps, where the main road and ramp run adjacent to each other, forming a small included angle. Moreover, solely relying on vehicle positioning and road geometry fails to solve the matching ambiguities regarding whether the vehicle is on elevated roads or underneath them.

In this letter, we present a new online map matching
method that can effectively address the above mentioned
challenges. Specifically, we leverage visual information including lane marking detections and scenario recognition in the conventional map matching method based on Hidden Markov Model (HMM).
The overall framework is illustrated in Fig. \ref{fig:2}, and the main contributions include:
\begin{figure}
    \centering
    \includegraphics[width=1\linewidth]{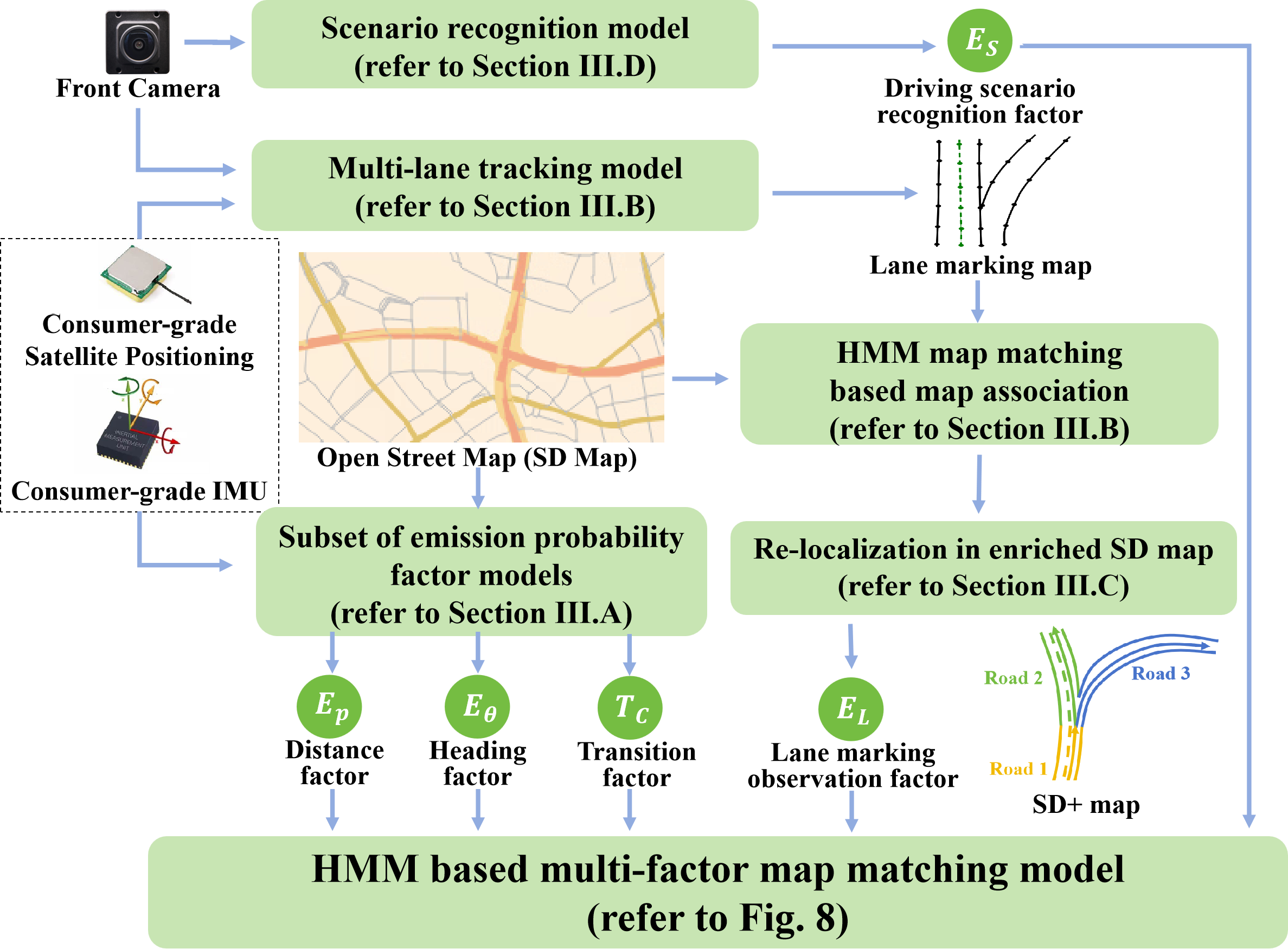}
    \caption{The overall framework of online map matching in complex road networks using lane markings and scenario recognition.}
    \label{fig:2}
\end{figure}
\begin{itemize}
\item \textbf{Online map matching using enriched SD map}: 
We use an multi-lane tracking method \cite{3} to generate the lane markings, which are then associated with the SD map, forming a lane-level enriched SD map. By leveraging lane marking information from the enriched SD map as probability factors in the HMM, our proposed method enables re-localization of the vehicle on the map and thereby enhances map matching accuracy at road splits.
\item \textbf{Integration of driving scenario recognition into HMM}: We incorporate the driving scenario recognition model by OMM-OBDSC \cite{4} into the design of the HMM factors for the online map matching method, enhancing map matching performance in multilevel road areas.
\item \textbf{Benchmark of online map matching in urban complex road network scenarios}: Extensive tests on the Zenseact Open Dataset \cite{22} and urban roads in Shanghai, China, covering complex road conditions, such as elevated roads, expressway ramps, and ordinary urban roads, show that our proposed method achieves $F_1$ scores of 98.04\% and 94.60\% on the Zenseact Open Dataset and test data of multilevel road
areas in Shanghai respectively, significantly outperforming benchmark methods.
\end{itemize}
\section{RELATED WORK}
Map matching has been an important research area, with some popular solutions proposed in the literature, e.g. the nearest neighbor matching from point to curve \cite{5} and from curve to curve \cite{6}, ST-match \cite{7}  and the HMM map matching algorithm \cite{8}. The HMM based map matching algorithm has been extensively studied due to its excellent performance. 
In addition, learning based map matching methods, such as DeepMM \cite{13}, L2MM \cite{15}, and GraphMM \cite{14} have emerged which model the vehicle trajectories and maps as the inputs of the neural network. However, the mainstream map matching methods solely rely on trajectory information without multi-sensor data integration, resulting in limited performance in complex and dense urban road networks.

 
In \cite{11}, lane-level map matching is performed with the assistance of surrounding lane markings (dashed, solid, etc.) using High Definition (HD) map. But given the limited coverage and high maintenance cost of HD maps, SD maps plays a crucial role for intelligent driving.
In \cite{zhang24:iros}, SD map and vehicle-mounted perception are used to generate online vectorized HD map representations. In Advanced Driver-Assistance Systems (ADAS), cameras that generate lane markings are being equipped on intelligent vehicles for localization \cite{association}. LaneMatch is proposed in \cite{lanematch} to localize vehicle by matching lane detection outputs with the lane-marking shapes and types extracted offline from satellite images. 

In addition to lane markings, road recognition can be also leveraged for map matching. In \cite{17}, specific combinations of filters for different operational environments are provided, which can adjust the matching process according to the traffic scenario. In \cite{amap},  an Elevated Road Network is proposed to address the elevated road recognition problem, which takes speed, satellite SNR and distance to elevated road as input for elevated road recognition. DSCMM \cite{20} and OMM-OBDSC \cite{4} used front-view images for driving scenario recognition to obtain road class probabilities by processing front-view images. 
In \cite{21} and \cite{eleUnit}, an Elevation-Aware Unit is proposed to utilize front-view images and IMU data to acquire elevation information for diverse urban roads. While these methods can mitigate errors from elevated roads and ordinary urban roads underneath them, they are less effective for road splits.

This letter presents an online map matching method based on scenario recognition and lane markings, aiming to enhance matching accuracy in complex road networks, especially in multilevel road areas.
\section{METHOD}
The proposed method extends the traditional HMM based map matching framework. In addition to conventional emission probability estimations, which are typically derived from the distance and included angle between the vehicle trajectory and candidate road segments, our approach leverages front-view camera images to generate lane markings and perform scenario recognition. By incorporating visual context, the method acquires richer semantic information, thereby improving the accuracy of the map matching process.



\subsection{HMM Map Matching Model} \label{sec:3a}
\subsubsection{State Space Model}
In the map matching model, the hidden state $x_N$ represents the actual road that the vehicle is on. For example, $x_N = r_i$ indicates that the vehicle is on road $r_i$ at time step $N$. The perception information, e.g. vehicle positions and front-view images, serve as the observations.

In HMM map matching model, it is assumed that the current state $x_N$ at time step $N$ depends only on the previous state $x_{N-1}$ at time step $N-1$, expressed as
\begin{equation}P(x_N|x_{N-1},x_{N-2},...,x_0)=P(x_N|x_{N-1}). 
\end{equation}

\subsubsection{Emission probability factor based on vehicle pose}
\begin{figure}
    \centering
    \includegraphics[width=0.65\linewidth]{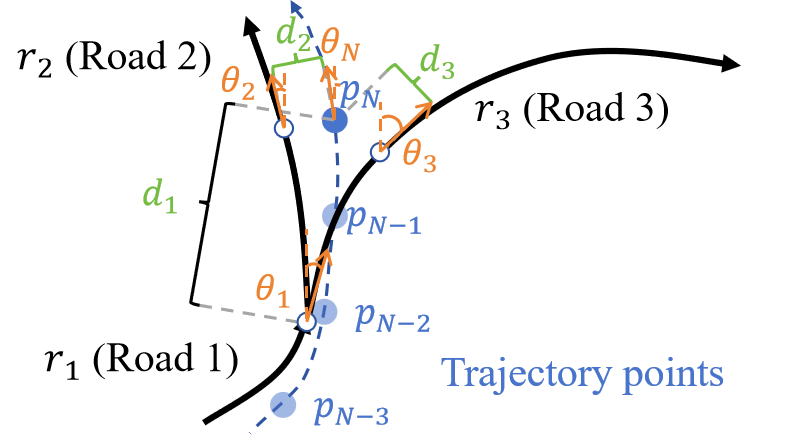}
    \caption{Schematic diagram of calculating the distance from the vehicle to the road and the road heading angle. In the diagram, $p_N$ represent vehicle positioning points at time $N$, which form the vehicle trajectory together with $p_{N - 1}$, $p_{N - 2}$, and $p_{N - 3}$. $\theta_N$ denotes the vehicle heading angle at time $N$. Specifically, $d_1$, $d_2$, and $d_3$ are the distances from the vehicle's positioning point $p_N$ to roads $r_1$, $r_2$, and $r_3$, respectively. When a projection point exists on the road, the perpendicular distance is taken; otherwise, the distance to the road endpoint is used. Additionally, $\theta_1$, $\theta_2$, and $\theta_3$ represent the road heading angles at the nearest points from the vehicle to roads $r_1$, $r_2$, and $r_3$, respectively.
}
    \label{fig:3}
\end{figure}
The main observations at time $N$ are the vehicle position $p_N$ and the vehicle heading angle $\theta_N$, as shown in Fig. \ref{fig:3}. Candidate roads within a certain area are queried based on the vehicle's current positioning, in the sense that the probability that the vehicle is on a given road decreases as the distance between the vehicle’s position and the road increases. We denote the distance between the vehicle position and the i-th candidate road $r_i$ using $d_i$, which is assumed to be zero-mean Gaussian distributed with standard deviation $\sigma$. Then the emission probability factor of distance $d_i$ is given by:
\begin{equation}
p_d(d_i|x_N=r_i)=\mathcal{N}(d_i;0,\sigma),
\end{equation}
where $\Delta\theta_i$ is the included angle of vehicle heading $\theta_N$ and road heading $\theta_i$ ($0^{\circ} \leq\theta_N,\theta_i\leq360^{\circ}$), with
\begin{equation}
\Delta\theta_i = \min\left(\left|\theta_i - \theta_N\right|,360^{\circ}-\left|\theta_i - \theta_N\right|\right),
\end{equation}
\begin{equation}
p_{\theta}(\Delta\theta_i|x_N = r_i) \propto
\begin{cases}
\frac{1 + \cos(2\Delta\theta_i)}{2}, & \text{if }  \Delta\theta_i < 90^\circ, 
\\
\varepsilon_1, & \text{otherwise}, 
\end{cases}
\label{eq:4}
\end{equation}
where $\varepsilon_1$ is the probability compensation value to deal with potential U-turn and reverse driving behaviors, and in this work it is set to $10^{-4}$. It can be seen from Eq. (\ref{eq:4}) that the probability of the vehicle being on candidate road $r_i$ decreases as $\Delta \theta_i$ increases. 
\subsubsection{Transition Probability Factor Based on Road Connectivity}
Roads with connectivity information can be regarded as a directed graph. The transition probability factor depends on the road connectivity, and to compute it we first divide the roads into three categories according to the connection with road $r_j$: the set $R_A$ consists of the successor roads directly connected to $r_j$ and also $r_j$ itself (i.e., \textit{j=i} ), whereas set $R_B$ consists of the roads connected to $r_j$ through other intermediate roads. The transition probability factor from road $r_j$ to road $r_i$ is then given by
\begin{equation}
P_T(x_N= r_i|x_{N-1} = r_j) \propto
\begin{cases}
1, & \text{if } r_i\in R_A ,
\\
e^{-\frac{l_{i,j}}{\gamma}}, & \text{if } r_i\in R_B,
\\
\varepsilon_2, & \text{otherwise},
\end{cases}
\end{equation}
where $l_{i,j}$ is the minimum connected distance between $r_i$ and $r_j$, $\gamma$ is the attenuation adjustment factor, and $\varepsilon_2$ is the probability compensation value, which is set to $10^{-4}$ to deal with potential road transitions that do not conform to traffic regulations.

\subsubsection{State Backtracking Using the Viterbi Algorithm}\label{sec:3a4}
We apply the Viterbi algorithm \cite{viterbi} for online dynamic programming to solve the optimal hidden state sequence of the Hidden Markov Model.
Here we use $H_N$ to represent the set of all candidate roads at time $N$. For each candidate road $r_i\in H_N$ of the state $x_N$ at time $N$, we use the Viterbi algorithm to calculate the joint probability and find the hidden state sequence with the maximum probability corresponding to each candidate road $r_i$. The joint probability $P(x_N = r_i,d_i, \Delta \theta_i| x_{N-1})$ of candidate road $r_i$ at time $N$ of the Hidden Markov Model is given by
\begin{equation}
\begin{split}
&P(x_N = r_i, d_i, \Delta \theta_i| x_{N - 1}) \\& \propto \max_{
r_j \in H_{N - 1}} \{ P(x_{N - 1} = r_j, d_j, \Delta \theta_j| x_{N - 2}) 
\\&~~~~~~~~~~~~~~~\times P_T(x_N = r_i|x_{N - 1} = r_j)
\\&~~~~~~~~~~~~~~~ \times p_d(d_i|x_N = r_i) \times p_{\theta}(\Delta\theta_i|x_N = r_i)
  \}.
\end{split}
\label{eq:6}
\end{equation}

After computing the unnormlized joint probabilities of $P(x_N = r_i, d_i, \Delta \theta_i| x_{N - 1})$ for each $r_i\in H_N$, these values need to be normalized to ensure that the joint probability Eq. (\ref{eq:6}) is a proper probability function. The optimal matched road $r_N^*$ at time $N$ is given by:
\begin{equation}
\begin{split}
r_N^*=\mathop{\arg\max}\limits_{r_i \in H_{N}}\ P(x_N = r_i, d_i, \Delta \theta_i| x_{N - 1}).
\end{split}
\end{equation}

By backtracking the state of the optimal matched road $r_N^*$ at time $N$ in the Hidden Markov Model, the maximum likelihood state sequence can be obtained, which is the road sequence matched to the vehicle trajectory.
\subsection{The Association of Generated Lane Markings with SD Map}\label{sec:3b}
In this work, we adopt the multi-lane tracking method proposed in \cite{3} to generate the lane markings. This method makes use of the lane marking detection points extracted from the camera data to continuously generate and track the lane markings as the vehicle travels. Note that generated lane markings is represented by a set of polylines, where each polyline is compactly parameterized by a sequence of B-spline control points. 

\begin{figure}
    \centering
    \includegraphics[width=1\linewidth]{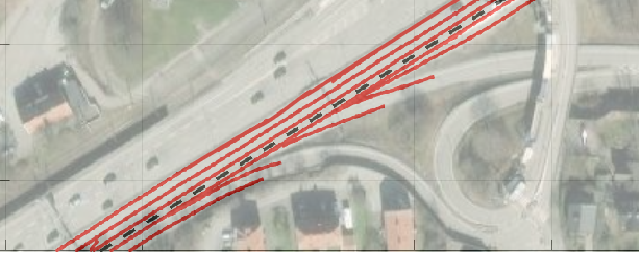}
    \caption{An example of the lane markings generated by the multi-lane tracking method. Each \textcolor[RGB]{255,0,0}{red curve} represents a tracked lane marking, and the \textbf{black dashed curve} represents the vehicle trajectory. The base map for reference is from Google Satellite Map.}
    \label{fig:4}
\end{figure}

An example of the generated lane markings is shown in Fig. \ref{fig:4}. With the lane markings in place, the map matching model described in Section \hyperref[sec:3a]{III.A} can be used to associate the tracked lane markings with the roads in the map.

Denote the set of tracked lane markings as $L=\{l_j|j=1,2,...,m\}$, where $m$ represent the total number of lane markings. The lane markings in the form of B-splines output by the multi-lane tracking method can be sampled to obtain discrete directed coordinate point sequences. Assume that each lane marking contains $n$ sampling points, denoted as 
$l_j=\{l_{j,1},l_{j,2},...,l_{j,n}\}$. 
Bring the coordinates $l_{j,k}$ and the heading angle $\theta_{j,k}$  of the sampled lane marking points into the model in Section \hyperref[sec:3a]{III.A}. The key is to adjust the value of the parameter $\sigma$ to fit the perpendicular distance distribution of the vehicle from the lane markings. 
We represent the association and subordination between $l_{j,k}$, $l_{j}$ and $r_i$ in the SD map in a simplified manner using a mapping function $\mathcal{M}_1(l_{j,k})=r_i$ and $\mathcal{M}_2(l_{j})=r_i$ respectively.  
Then based on the map matching algorithm, $P_m(\mathcal{M}_1(l_{j,k})=r_i)$, the probability of each tracked lane marking sampling point $l_{j,k}$ being matched to each candidate road $r_i$ can be calculated. In addition, by tracing back from the optimal matching sequence, the probability of each sampled point of the lane markings being matched to other roads can be obtained. The maximum value of these probabilities of being matched to each candidate road as the association probability between $l_j$ and $r_i$ can be obtained as
\begin{equation}
\begin{split}
P(\mathcal{M}_2(l_{j})=r_i)=\max_{k=1,2,...,n}P(\mathcal{M}_1(l_{j,k})= r_i) .
\end{split}
\end{equation}

\begin{figure}
    \centering
    \begin{subfigure}[t]{0.237\textwidth}
        \includegraphics[width=\textwidth]{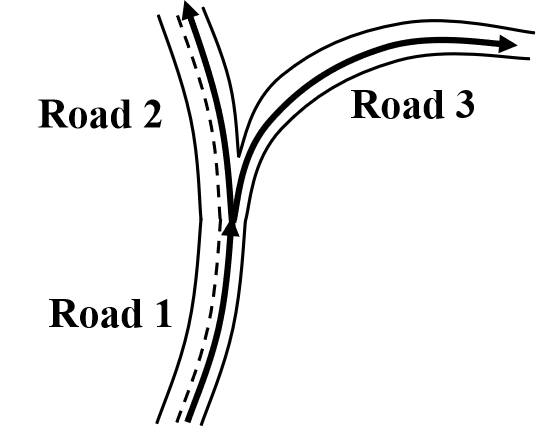}
        \subcaption{Before association}\label{fig:sub-5a}
    \end{subfigure}
    \hfill
    \begin{subfigure}[t]{0.237\textwidth}
        \includegraphics[width=\textwidth]{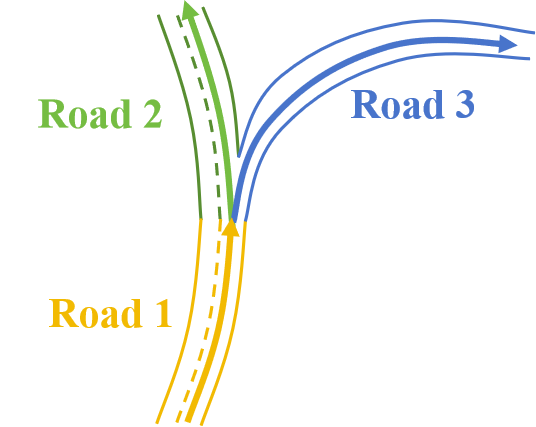}
        \subcaption{After association}\label{fig:sub-5b}
    \end{subfigure}
    \caption{Schematic diagram of enriched SD map fusion. The \textbf{thick polylines}with arrows are the roads of SD map, and the thin solid and dashed polylines represent the solid and dashed lane markings, respectively. In this figure, (a) shows the overlay of lane markings and the SD map in the same coordinate system, and note that there is no correlation between the lane markings and the SD map before performing the association. (b) shows the association result of multiple lane markings with multiple roads. The associated roads and lane markings are marked with the same color In this figure. One road can be associated with multiple lane marking instances, and different parts of one lane marking instance can also be associated with multiple roads with different probabilities. }
    \label{fig:5}
\end{figure}

An example of the association between the lanes and the roads is shown in Fig. \ref{fig:5}.
Through associating the lane markings with the roads, a lane-level enriched SD map is formed, and therefore the online map matching of the vehicle passing through the mapped area can be assisted by better map information.
\subsection{Emission Probability Factor of Lane Marking Detection}
\begin{figure}
    \centering
    \begin{subfigure}[t]{0.234\textwidth}
        \includegraphics[width=\textwidth]{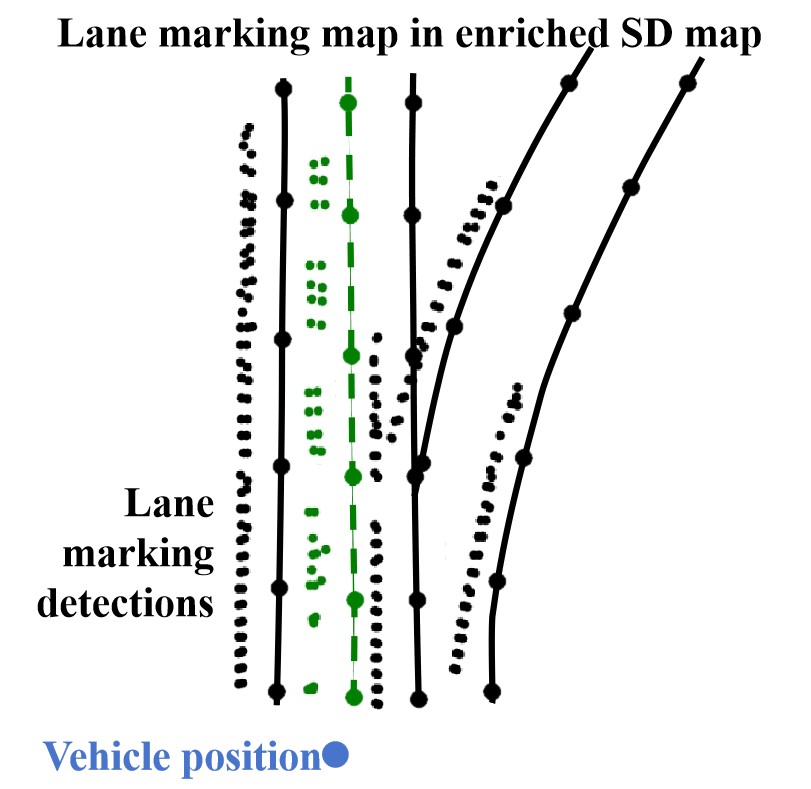}
        \subcaption{Before registration}\label{fig:sub-6a}
    \end{subfigure}
    \hfill
    \begin{subfigure}[t]{0.234\textwidth}
        \includegraphics[width=\textwidth]{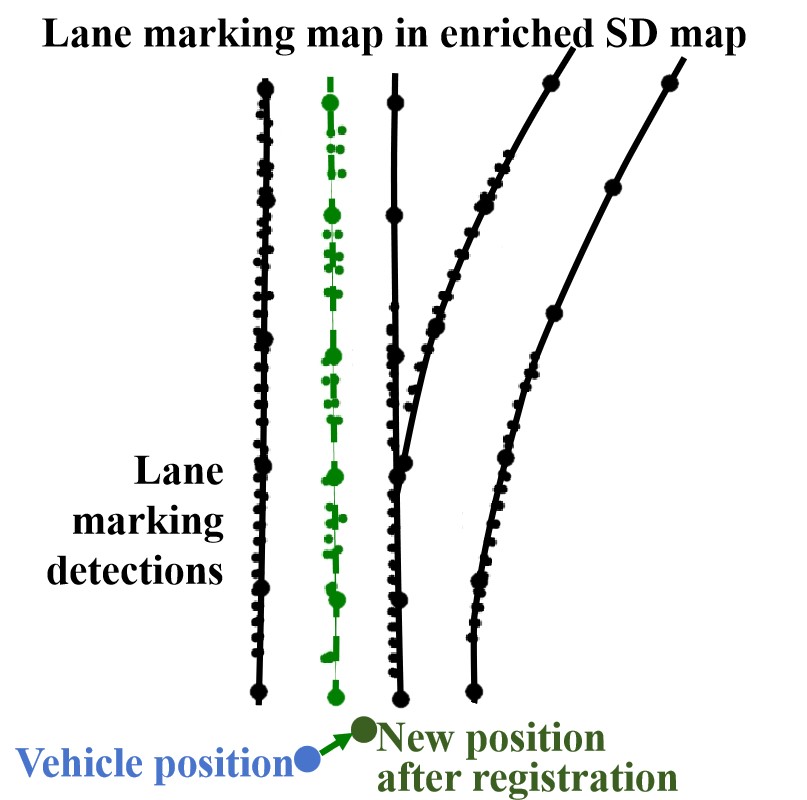}
        \subcaption{After registration}\label{fig:sub-6b}
    \end{subfigure}
    \caption{The process of lane marking ICP registration. The points in this figure are from the lane marking detection extracted from the front-view camera, and the polylines are from the enriched SD map. The \textcolor[RGB]{0,128,0}{green parts} represent the dashed lanes, and the \textbf{black parts} represent the solid lanes. In this figure, (a) shows the lane marking detection point cloud and the lane markings from enriched SD map  before registration. (b) shows the lane marking detection point cloud and the lane markings from enriched SD map after registration. It can be seen that after performing ICP registration, the vehicle can re-localize itself in enriched SD map.}
    \label{fig:6}
\end{figure}

Since there are positioning biases in each vehicle travel, in order to judge the current lane where the vehicle is located and obtain the information of the surrounding lane markings, we adopt the Iterative Closest Point (ICP) algorithm \cite{icp} to locate the vehicle in the lane-level enriched SD map.
The registration loss $L_{r}$ between the vehicle lane marking detection point and the points in the enriched SD map during the nearest neighbor point query is given by
\begin{align}
L_{r}(p_v,p_m) &= \sqrt{(x_v - x_m)^2+(y_v - y_m)^2+L_{type}^2}, \label{eq1} \\
L_{type} &= \begin{cases}
0, & \text{if }t_v = t_m, \\
f_{type}, & \text{if } t_v \neq t_m,
\end{cases} \label{eq2}
\end{align}
where $t_v$ and $t_m$ are, respectively, the lane marking type of the vehicle lane marking detection point and the lane marking type of the down-sampled point in the enriched SD map; $L_{type}$ is the registration loss of the lane marking type, and $L_{type} = f_{type}$ if $t_v \neq t_m$. A larger value of $f_{type}$ indicates a greater confidence in the classification accuracy of the lane marking detection algorithm. A comparison of vehicle position before and after registration is shown in Fig. \ref{fig:6}.

After the lane markings were associated with the SD map in Section \hyperref[sec:3b]{III.B}, an enriched SD map was formed. By using the positions and included angles of all the lane markings $L_N$ near the position of the vehicle at time $N$, the projection points of the vehicle position onto the adjacent lane markings are calculated, and the distance $d_{lane,j}$ between the vehicle and the lane markings as well as the heading included angle  $\Delta\theta_j=|\theta_{lane,j}-\theta_N|$  are obtained. Then, combined with the association probability between the lane markings and the roads, the lane marking detection probability factor of the vehicle for each candidate road can be calculated as
\begin{equation}
\begin{split}
P_L(x_N=r_i&|L_N) \propto \sum\limits_{l_j\in L_N} P_m(\mathcal{M}_2(l_{j})=r_i))\\
&\times \mathcal{N}(d_i;0,\sigma)\times p_{\theta}(\Delta\theta_j|x_N = r_i)  .
\end{split}
\end{equation}
\subsection{Emission Probability Factor of Driving Scenario Recognition}
\begin{figure}
    \centering
    \includegraphics[width=1\linewidth]{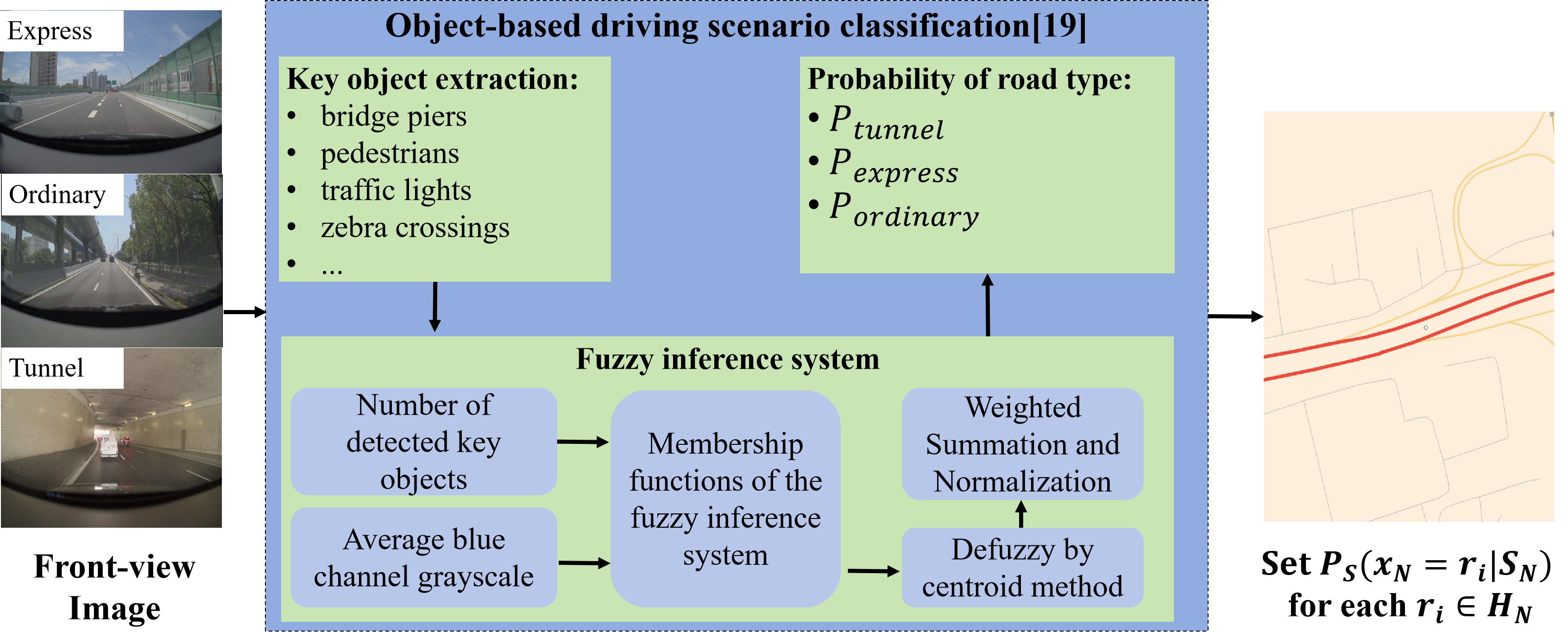}
    \caption{The process of deriving driving scenario recognition probability factors from the front-view image. The driving scenario recognition model from OMM-OBDSC \cite{4} is used to process the front-view image to obtain the probabilities of the vehicle being on the different roads. The method first uses YOLO object detector \cite{yolo}, which has been trained to detect the typical objects of different road types, including bridge piers, pedestrians, traffic lights, zebra crossings, non-motor vehicles, etc. Next, the average grayscale value of the blue channel of the image is calculated as an input of the fussy inference system. In the end, the emission probability factors of driving scenario recognition are generated for each candidate road.}
    \label{fig:7}
\end{figure}

In the complex urban road network, different types of roads are closely adjacent in the three-dimensional space. This makes it difficult for the vehicle to effectively distinguish the actual matched road. To address this problem, we make use of driving scenario recognition in the calculation of the emission probability factor. In this work, we use the driving scenario recognition model from OMM-OBDSC \cite{4} to process the images from the front-view camera. And by doing so we can obtain the probabilities that the scenario belong to ordinary urban roads, expressways, and tunnels, denoted using $P_{ordinary}$, $P_{express}$, $P_{tunnel}$, respectively as shown in Fig. \ref{fig:7}. We refer the readers to \cite{19} for more details of the driving scenario recognition module.
Then, the output of the above OBDSC model is added as a probability factor to the Hidden Markov Model, which is given by
\begin{equation}
\begin{split}
P_S(x_N=r_i|S_N) =\begin{cases}
P_{ordinary}, & \text{if }r_i \in R_N ,
\\
P_{express}, & \text{if }r_i \in R_E ,
\\P_{tunnel}, & \text{if }r_i \in R_T,
\\
\end{cases}
\end{split}
\end{equation}
where $S_N$ is the overall scenario information captured from the front-view image, $R_N$ is the set consisting of ordinary urban roads in the map, $R_E$ is the set consisting of elevated road or expressways in the map, and $R_T$ is the set consisting of tunnel roads.

\subsection{Improved Hidden Markov Model Probability Factor Graph}
\begin{figure}
    \centering
    \includegraphics[width=1\linewidth]{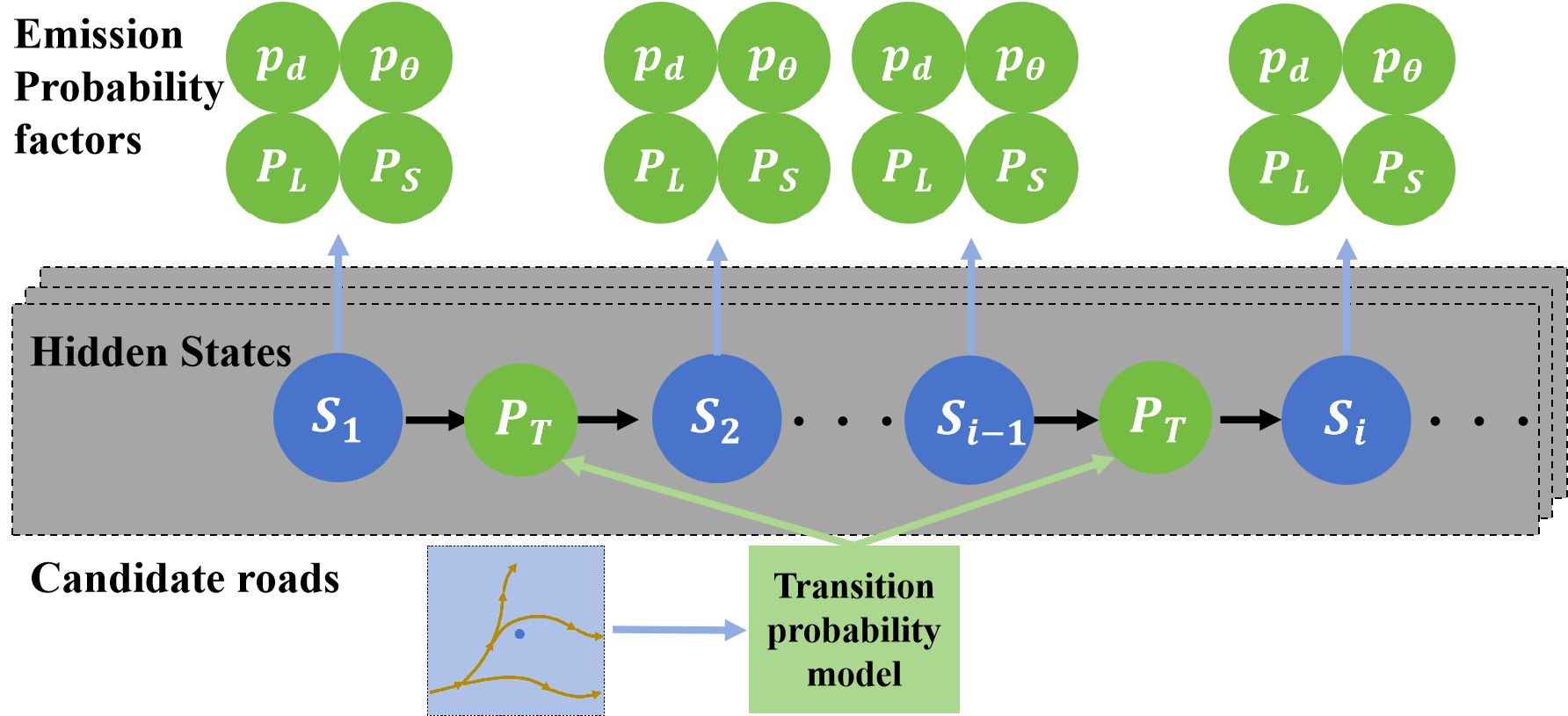}
    \caption{HMM with multiple probability factors.}
    \label{fig:8}
\end{figure}

After the incorporation of enriched SD map and driving scenario recognition, the complete Hidden Markov Model with multiple probability factors is shown in Fig. \ref{fig:8}.
The joint probability of the state $x_N$ at time $N$ of the Hidden Markov Model can be finally expressed as
\begin{equation}
\begin{split}
&P(x_N = r_i,d_i, \Delta \theta_i| x_{N-1})\\& \propto \max_{r_j \in H_{N-1}}\{  P(x_{N-1} = r_j,d_j, \Delta \theta_j|  x_{N-2})\\
&~~~~~~~~~~~~~~~\times P_T(x_N = r_i|x_{N-1} = r_j)\\
&~~~~~~~~~~~~~~~ \times  p_d(d_i|x_N = r_i) \times p_{\theta}(\Delta\theta_i|x_N = r_i)
  \\
& ~~~~~~~~~~~~~~~\times P_L(x_N=r_i|L_N) \times P_S(x_N=r_i|S_N) \}.
\end{split}
\end{equation}
Then the Viterbi algorithm mentioned in Section \hyperref[sec:3a4]{III.A} can be used to find the maximum likelihood state sequence, which are the matched roads corresponding to the vehicle trajectory.
\section{EXPERIMENTS}

\begin{figure}
    \centering
    \includegraphics[width=1\linewidth]{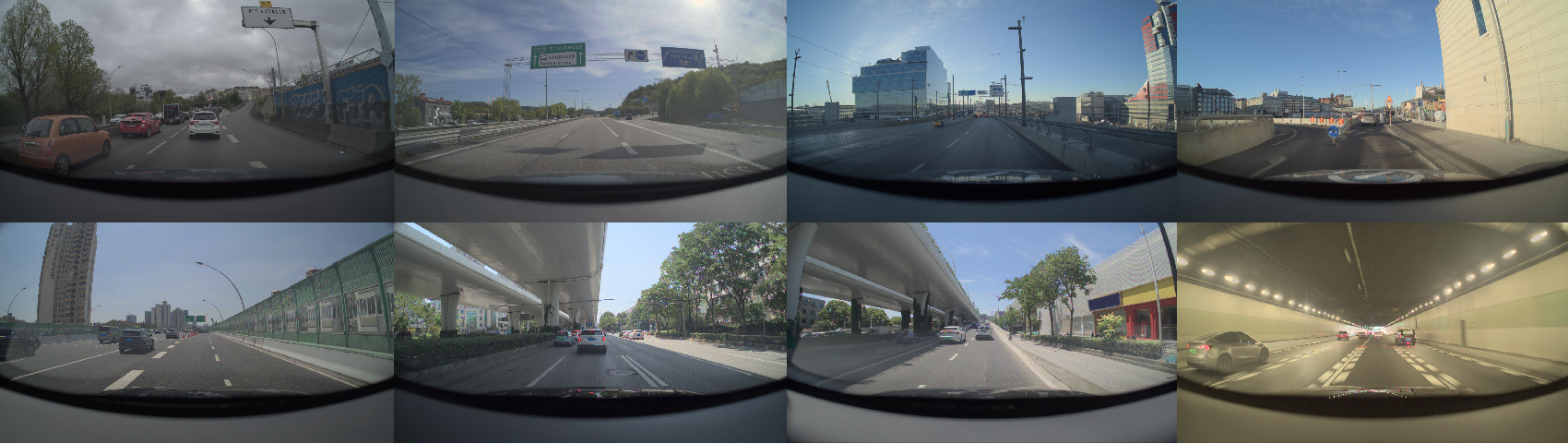}
    \caption{An overview of the driving scenarios for testing the map matching algorithm.}
    \label{fig:9}
\end{figure}
\newcolumntype{L}[1]{>{\raggedright\arraybackslash}p{#1}}
\newcolumntype{C}{>{\centering\arraybackslash}X} 
\begin{table*}
    \caption{Results on Drives of Zenseact Open Dataset \cite{22}}
    \label{tab:1}
    \centering
        \begin{ThreePartTable}
    \begin{tabularx}{\textwidth}{L{0.21\textwidth} L{0.15\textwidth} C C C C}
        \toprule
        Method & Positioning inputs \cite{22} & \makecell{$\textit{MatchRate}$\\(\%)}$\uparrow$ & \makecell{$\textit{Precision}$\\(\%)}$\uparrow$ & \makecell{$\textit{Recall}$\\(\%)}$\uparrow$ & \makecell{$F_1$ score\\(\%)}$\uparrow$ \\
        \midrule
        Nearest neighbor \cite{5} & GNSS & 91.43 & 89.06 & 89.43 & 89.25 \\
        Nearest neighbor \cite{5} & GNSS+IMU  & 92.48 & 90.10 & 90.51 & 90.30 \\
        Nearest neighbor \cite{5} & OxTS RT3000 & 95.10 & 93.49 & 94.10 & 93.80 \\
        Online HMM \cite{8} & GNSS & 96.51 & 95.30 & 95.41 & 93.73 \\
        AMM \cite{24} & GNSS & 94.39 & 93.88 & 94.07 & 93.97 \\
        Online HMM $+$ OBDSC \cite{4} & GNSS & 96.97 & 96.08 & 96.18 & 96.13 \\
        Proposed method & GNSS & 98.26 & 97.83 & 97.92 & 97.87 \\
        Proposed method & GNSS+IMU & \textbf{98.35} & \textbf{98.00} & \textbf{98.08} & \textbf{98.04} \\
        \bottomrule
    \end{tabularx}
\begin{tablenotes}
\item[$\uparrow$]The upper arrow denotes that better performance is registered with higher value.
\end{tablenotes}
    \end{ThreePartTable}
\end{table*}

\begin{table*}
    \caption{Results on Test Data of Multilevel Road Areas in Shanghai}
    \label{tab:2}
    \centering
    \begin{tabularx}{\textwidth}{L{0.21\textwidth} L{0.15\textwidth} C C C C}
        \toprule
        Method & Positioning inputs \cite{22} & \makecell{$\textit{MatchRate}$\\(\%)}$\uparrow$ & \makecell{$\textit{Precision}$\\(\%)}$\uparrow$ & \makecell{$\textit{Recall}$\\(\%)}$\uparrow$ & \makecell{$F_1$ score\\(\%)}$\uparrow$ \\
        \midrule
        Nearest neighbor \cite{5} & GNSS & 64.53 & 43.10 & 43.81 & 43.45 \\
        Nearest neighbor \cite{5} & GNSS$+$IMU & 65.14 & 44.65 & 45.68 & 45.16 \\
        Nearest neighbor \cite{5} & OxTS RT3000 & 78.09 & 65.05 & 65.46 & 65.25 \\
        Online HMM \cite{8} & GNSS & 74.56 & 62.90 & 63.78 & 63.33 \\
        AMM \cite{24} & GNSS & 67.47 & 66.84 & 66.89 & 66.87 \\
        Online HMM $+$ OBDSC \cite{4} & GNSS & 89.63 & 88.23 & 89.27 & 88.75 \\
        Proposed method & GNSS & 91.60 & 93.85 & 94.57 & 94.21 \\
        Proposed method & GNSS+IMU & \textbf{92.95} & \textbf{93.99} & \textbf{95.21} & \textbf{94.60} \\
        \bottomrule
    \end{tabularx}
\end{table*}
The SD map used in this work is OpenStreetMap (OSM) \cite{osm}. 
Subsequent tests are carried out on Drives of Zenseact Open Dataset \cite{22} and road tests covering 102.6 kilometers in Shanghai, China. The test scenarios cover various types of roads ranging from expressways to rural roads. Examples of these scenarios are shown in Fig. \ref{fig:9}.
\subsection{Map Matching Evaluation Metrics}

In this work we use $\textit{MatchRate}$, $\textit{Precision}$, $\textit{Recall}$ and $F_1$ score for performance evaluation.

The $\textit{MatchRate}$ is defined as the proportion of correctly matched trajectory points $N_{correct}$ among all trajectory points $N_{all}$:
\begin{equation}
MatchRate=\frac{N_{correct}}{N_{all}},
\end{equation}

We let $\textit{TP}$ be the true positives, $\textit{FP}$ be the false positives, $\textit{FN}$ be the false negatives. Then the $\textit{Precision}$ is the fraction of relevant instances among the retrieved instances, expressed as $Precision=\frac{TP}{TP+FP}$. The $\textit{Recall}$ is the fraction of relevant instances that have been retrieved over the total amount of relevant instances, expressed as $Recall=\frac{TP}{TP+FN}$. In the context of map matching, the $\textit{Precision}$ and $\textit{Recall}$ can be, respectively, defined as
\begin{equation}
Precision=\frac{L_{correct}}{L_{mm}}.
\end{equation}
\begin{equation}
Recall=\frac{L_{correct}}{L_{gt}}.
\end{equation}
where $L_{gt}$ is the length of the ground truth trajectory, $L_{mm}$ is the length of the map matching trajectory, and $L_{correct}$ is the length of the correct overlapping road segments between the map matching trajectory and ground truth trajectory. 

The $F_1$ score is defined using $\textit{Precision}$ and $\textit{Recall}$ as
\begin{equation}
F_1 =\frac{2\times Precision\times Recall}{Precision+Recall}.
\end{equation}

\subsection{Test on Drives of Zenseact Open Dataset}
In order to evaluate the effectiveness of the proposed method in the online navigation and positioning function, we conducted tests on Drives of Zenseact Open Dataset, which covers various road types in Sweden and France. 
The comparison of results between our proposed method and other methods is shown in Table \ref{tab:1}. First we apply nearest neighbor matching (point-to-curve \cite{5}) using three different positioning inputs: standalone GNSS, fused GNSS/IMU, and the high-precision OxTS RT3000 (refer to \cite{22} for device specifications). All positioning data were down-sampled to 1 FPS to ensure consistency with GNSS sampling rates. When used as inputs for map matching, the $F_1$ score improved by 1.05\% and 4.55\%, respectively. However, these improvements remain insufficient to reach the performance level of mainstream map matching algorithms. This suggests that enhancing positioning accuracy alone is inadequate to fully address the challenges of map matching.
Consequently, for a more comprehensive comparison, we evaluate our proposed method against four other benchmark methods. According to the experimental results, our proposed method achieved an average $\textit{MatchRate}$, $\textit{Precision}$, $\textit{Recall}$, $F_1$ score of 98.35\%, 98.00\%, 98.08\% and 98.04\%, respectively. All evaluation metrics are superior to those of the comparative methods.

In order to further test the performance of our proposed method in more complex road networks, we have conducted experiment on test data of multilevel road areas in Shanghai.
\begin{table*}
    \caption{Ablation Experiments on Both Datasets} 
    \label{tab:3}
    \centering
    \begin{ThreePartTable}
    \begin{tabularx}{\textwidth}{L{0.20\textwidth} L{0.23\textwidth} C C C C}
        \toprule
        Dataset & Model & \makecell{$\textit{MatchRate}$\\(\%)}$\uparrow$ & \makecell{$\textit{Precision}$\\(\%)} $\uparrow$& \makecell{$\textit{Recall}$\\(\%)}$\uparrow$ & \makecell{$F_1$ score\\(\%)}$\uparrow$ \\
        \midrule
        \multirow{4}{*}{ \makecell{Zenseact Open Dataset \cite{22}\\(Drives)}} 
            & \textcolor[RGB]{238,130,47}{$P_L$ and $P_S$ ablated (Baseline)} & 97.58 & 96.72 & 96.83 & 96.77 \\
            & $+P_S$\tnote{1} & 97.95 (+0.37) & 97.28 (+0.56) & 97.38 (+0.55) & 97.33 (+0.56) \\
            & $+P_L$\tnote{2} & 98.16 (+0.58) & 97.69 (+0.97) & 97.79 (+0.96) & 97.74 (+0.97) \\
            &\textcolor[RGB]{117,189,66}{$+P_S+P_L$(Proposed method)} & \textbf{98.26 (+0.68)} & \textbf{97.83 (+1.11)} & \textbf{97.92 (+1.09)} & \textbf{97.87 (+1.10)} \\
        \midrule
        \multirow{4}{*}{\makecell{Test data of multilevel road \\ 
        areas in Shanghai}} 
            & \textcolor[RGB]{238,130,47}{$P_L$ and $P_S$ ablated (Baseline)} & 75.41 & 65.89 & 66.75 & 66.32 \\
            & $+P_S$ & 90.38 (+14.97) & 92.31 (+26.42) & 93.18 (+26.43) & 92.74 (+26.42) \\
            & $+P_L$ & 82.95 (+7.54) & 78.87 (+12.98) & 79.58 (+12.83) & 79.22 (+12.90) \\
            &\textcolor[RGB]{117,189,66}{$+P_S+P_L$(Proposed method)} & \textbf{91.60 (+16.19)} & \textbf{93.85 (+27.96)} & \textbf{94.57 (+27.82)} & \textbf{94.21 (+27.89)} \\
        \bottomrule
    \end{tabularx}
    \begin{tablenotes}
     \item[1]Emission probability factor of driving scenario recognition \item[2]Emission probability factor of lane marking detection   
    \end{tablenotes}

    \end{ThreePartTable}
\end{table*}

\subsection{Test on Multilevel Road Areas in Shanghai}
\begin{figure}
    \centering
    \includegraphics[width=0.8\linewidth]{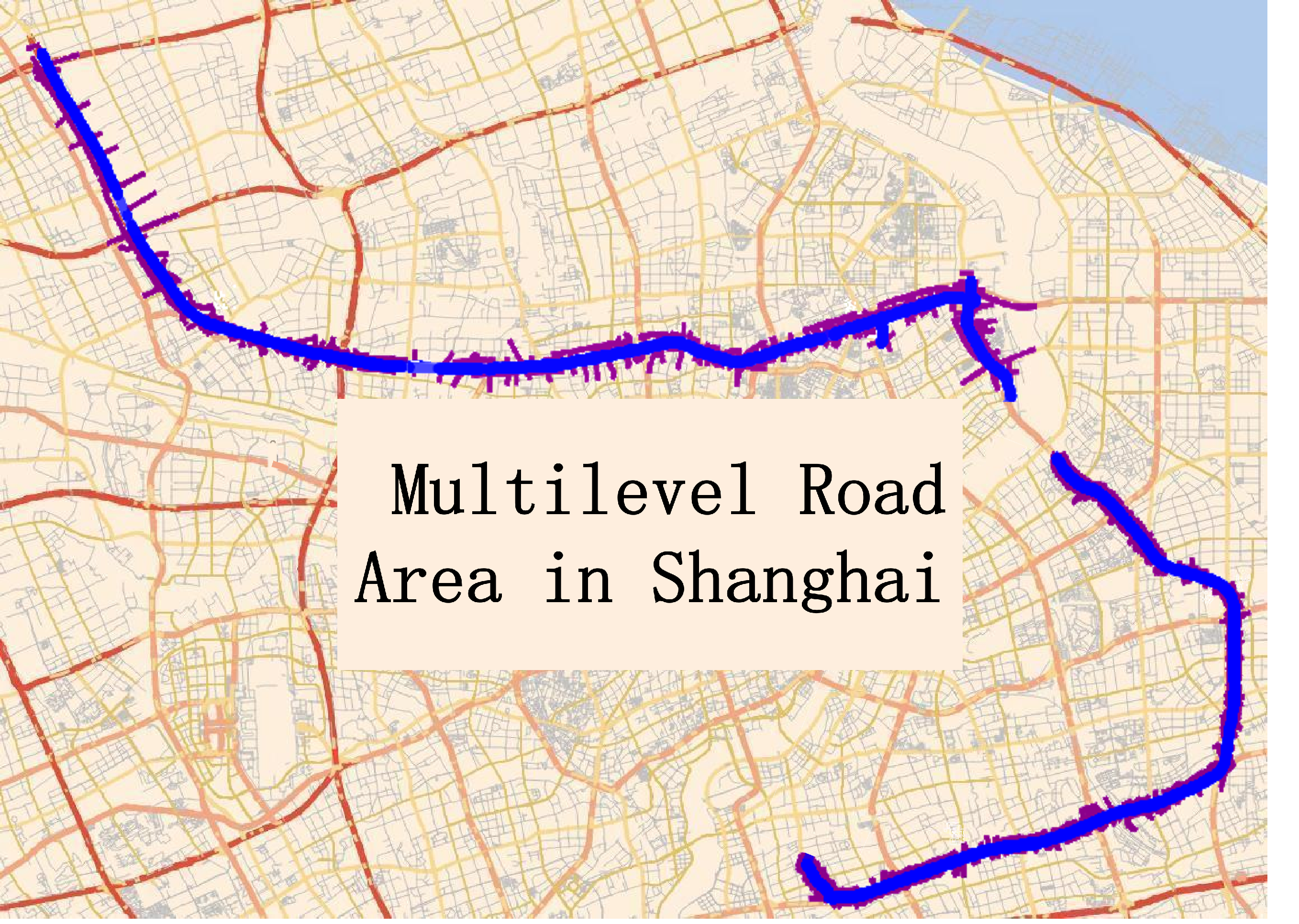}
    \caption{The test trajectories are shown on the SD map (OpenStreetMap\cite{osm}) within the multilevel road areas in Shanghai. The \textcolor[RGB]{0,0,255}{blue polylines} represent the trajectories.
    }
    \label{fig:10}
\end{figure}

We conducted the data collection experiment using a vehicle with the same sensor configuration as that of the Zenseact Open Dataset \cite{22}. The trajectory in China covers the multilevel road areas in Shanghai, which includes different road types such as road splits in expressway, the elevated roads and ordinary urban roads underneath them, and tunnels. The trajectory of the tested scenario is shown in Fig. \ref{fig:10}. Further verification has been carried out on the driving data totaling 8680 seconds in duration and covering an accumulated distance of 102.6 kilometers. This is used to verify the effectiveness of the method proposed in this letter for matching in complex urban road areas. The test results are shown in Table \ref{tab:2}. 
The online map matching task in Shanghai is more challenging due to the city's complex, dense, and multi-level road network, compared to the driving data from the Zenseact Open Dataset, which was captured in Europe.

Since the elevated roads and ordinary urban roads underneath them are parallel and adjacent in Shanghai multilevel road area, the vehicles can easily fail to distinguish whether they have entered or left the elevated road through the ramp. Therefore, the emission probability factor of driving scenario recognition $P_S$ and the emission probability factor of lane marking detection $P_L$ play a very significant role in the map matching algorithm. Notably, they can prevent the vehicle from being wrongly matched to the surface road while driving on the elevated road, or vice versa. 
The experiments show that our proposed method outperforms the benchmarking methods by a large margin on test data of multilevel road areas in Shanghai.

\begin{figure}[htbp]
    \centering
    \includegraphics[width=0.48\textwidth]{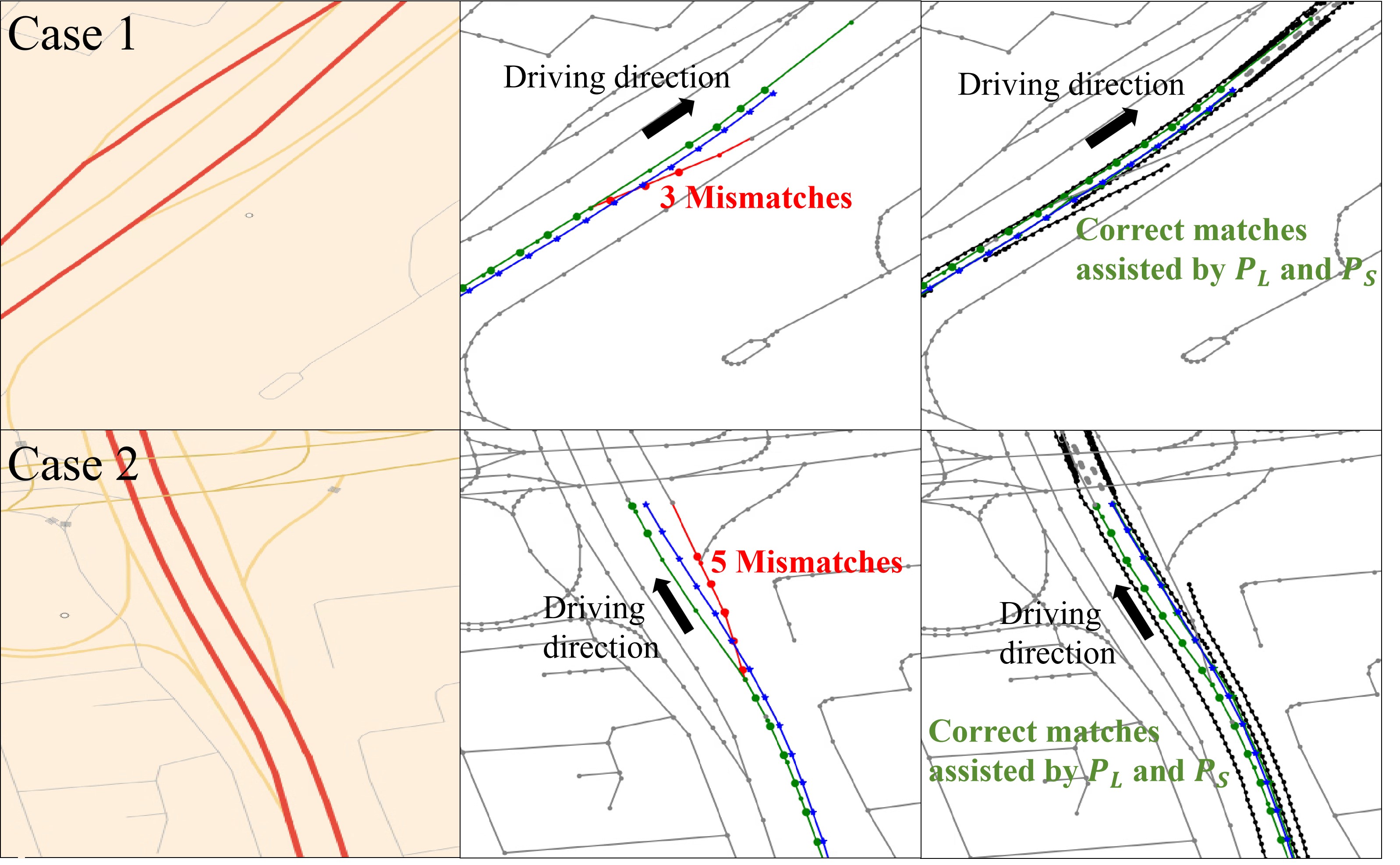}
    \begin{subfigure}[t]{0.157\textwidth}
        \subcaption{SD Map colored by different road class}\label{fig:sub-11a}
    \end{subfigure}
    \hfill
    \begin{subfigure}[t]{0.157\textwidth}
        \subcaption{\textcolor[RGB]{238,130,47}{Baseline in Table III}}\label{fig:sub-11b}
    \end{subfigure}
    \hfill
    \begin{subfigure}[t]{0.157\textwidth}
        \subcaption{\textcolor[RGB]{117,189,66}{Proposed method}}\label{fig:sub-11c}
    \end{subfigure}

    \caption{
    Improvements of online map matching in the case of road splits. Our proposed method enables vehicles to be localized on the correct roads with the aid of enriched SD map, so that addresses the issue of incorrect map matching due to trajectory deviations.
    In (a), the SD map utilized by the map matching algorithm is visualized. 
    In (b), the matching errors that occurred after ablating $P_S$ and $P_L$ are illustrated. The \textcolor[RGB]{150,150,150}{solid gray curve} represents the roads in the SD map. The \textcolor[RGB]{0,0,255}{solid blue polyline} depict the vehicle's trajectory, while the \textcolor[RGB]{0,0,255}{blue stars} indicate discrete trajectory points. The \textcolor[RGB]{0,255,0}{solid green polylines} denote the correctly matched roads, and the \textcolor[RGB]{0,255,0}{green circles} represent correct road projection points. Conversely, the \textcolor[RGB]{255,0,0}{solid red polylines} illustrate the incorrectly matched roads, along with the \textcolor[RGB]{255,0,0}{red circles} indicating incorrect road projection points. In (c), the accurate matching results generated by our proposed method are shown. The enriched SD map is shown, and both $P_S$ and $P_L$ are employed in the map matching process. The \textbf{solid black polylines} correspond to solid lane markings, while the \textcolor[RGB]{150,150,150}{gray dashed polylines} represent dashed lane markings. }
    \label{fig:11}
\end{figure}

\begin{figure}[htbp]
    \centering
    \includegraphics[width=0.48\textwidth]{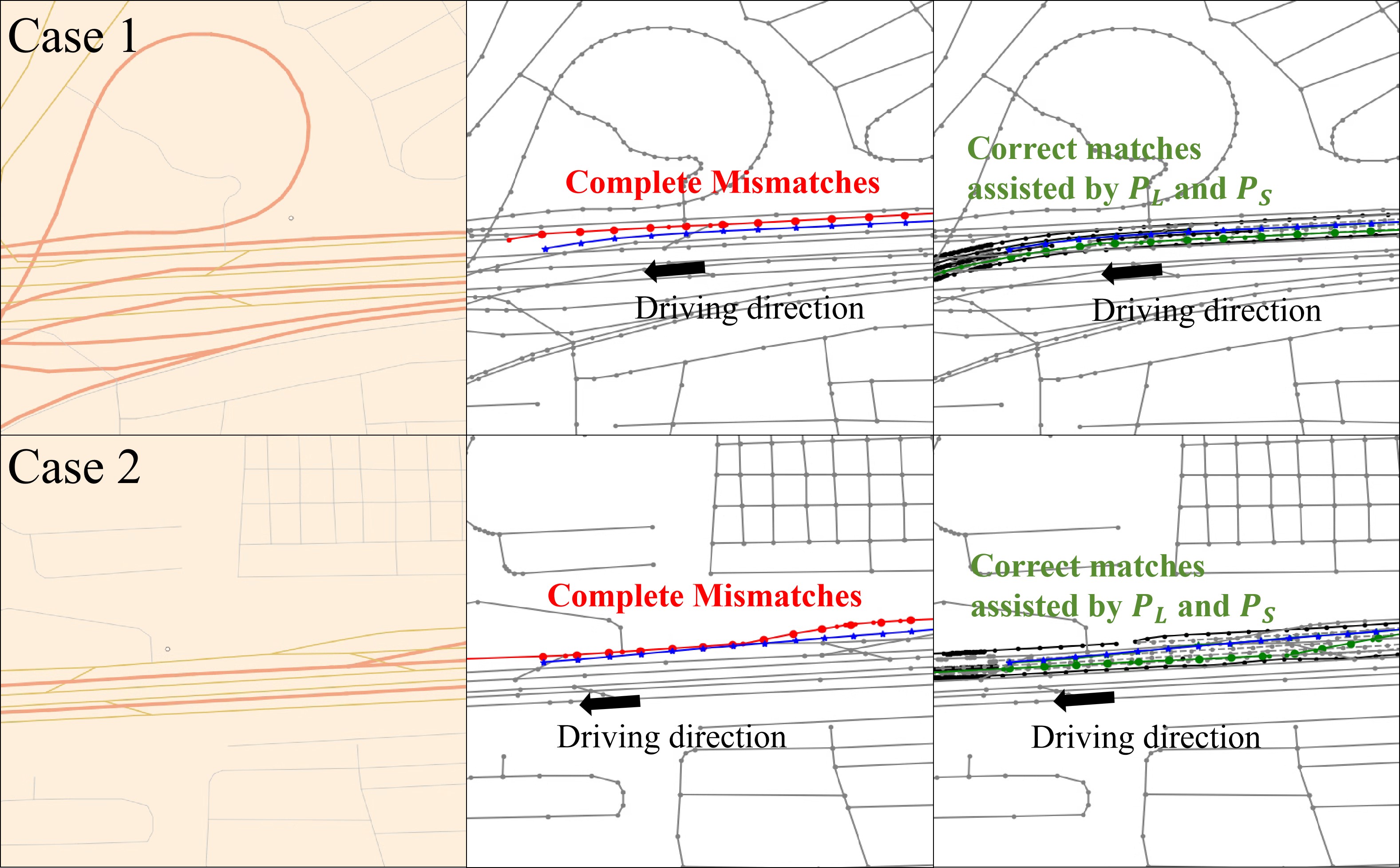}
    \begin{subfigure}[t]{0.157\textwidth}
        \subcaption{SD Map colored by different road class}\label{fig:sub-12a}
    \end{subfigure}
    \hfill
    \begin{subfigure}[t]{0.157\textwidth}
        \subcaption{\textcolor[RGB]{238,130,47}{Baseline in Table III}}\label{fig:sub-12b}
    \end{subfigure}
    \hfill
    \begin{subfigure}[t]{0.157\textwidth}
        \subcaption{\textcolor[RGB]{117,189,66}{Proposed method}}\label{fig:sub-12c}
    \end{subfigure}

    \caption{Improvements of online map matching in the case of multilevel road areas. Our proposed method effectively enables vehicles to be localized on the correct elevated roads with the aid of driving scenario recognition.
}
    \label{fig:12}
\end{figure}

\subsection{Ablation study}
To more precisely assess the contributions of each module to the evaluation metrics, we carried out ablation experiments relying solely on GNSS to ensure a fair comparison. The outcomes of these experiments are presented in Table \ref{tab:3}.

An examination of the ablation experiment results in Table \ref{tab:3} reveals that both the emission probability factor of driving scenario recognition, denoted as $P_S$, and the emission probability factor of lane marking detection, denoted as $P_L$, have a positive impact on the matching evaluation metrics. In the driving scenarios of the Zenseact Open Dataset, $P_L$ plays a more significant role. Incorporating $P_L$ can effectively enhance the evaluation metrics, particularly when the vehicle is required to differentiate the correct road at road junctions.

In the tests conducted in the multilevel road areas of Shanghai, the emission probability factor of driving scenario recognition, $P_S$, plays a more prominent role. When both of these factors are removed from the method, the $\textit{MatchRate}$ drops to 75.41\%, and the $F_1$ score drops to 66.32\%. When only $P_S$ is introduced, the $F_1$ score experiences a 26.42\% increase, reaching 92.74\%. Conversely, when only $P_L$ is added,  and the $F_1$ score rises to 79.22\%. However, both of these values are lower than the $F_1$ score of 94.21\% achieved by the proposed method before ablation.

Fig. \ref{fig:11} and Fig. \ref{fig:12} shows the improvements of our proposed method in several scenarios. This comparison underscores the significance of the emission probability factor of driving scenario recognition $P_S$ and the emission probability factor of lane marking detection $P_L$ in distinguishing between elevated roads and ordinary urban roads, as well as in differentiating the split roads of expressways.    

\section{CONCLUSIONS}

In this work, we propose a HMM-based multi-factor online map matching method that integrates lane markings to form the enriched SD map and driving scenario recognition using on-board camera input. By incorporating visual context, the method enhances semantic understanding of the driving environment, leading to improved accuracy in online map matching. Extensive evaluations on the Zenseact Open Dataset and in complex road networks in Shanghai demonstrate that our approach outperforms existing methods, particularly in challenging scenarios such as road splits and multilevel road areas.

\section*{Acknowledgement}
The authors from Tongji University were funded in part by the National Key R\&D Program of China under Grant 2022YFE0117100. This work was completed through the collaboration between Tongji University and Zenseact. The cooperation brought together complementary expertise and resources, facilitating the smooth progress of our research. We are deeply indebted to both institutions for their contributions to this work.

\printbibliography

\end{document}